\RequirePackage{fix-cm}
\documentclass{article}          
\usepackage{graphicx}
\usepackage{multirow}
\usepackage[table,xcdraw]{xcolor}
\usepackage{amsmath}
\usepackage{amsfonts}
\title{Deep reinforcement learning applied to an assembly sequence planning problem with user preferences
	\thanks{\textit{\underline{Citation}}: 
		\textbf{Neves, M., Neto, P. Deep reinforcement learning applied to an assembly sequence planning problem with user preferences. Int J Adv Manuf Technol 122, 4235–4245 (2022). DOI: 10.1007/s00170-022-09877-8}} 
}

\author{
	Miguel Neves, Pedro Neto \\
	University of Coimbra \\
	Coimbra\\
}

\begin{document}
	\maketitle


\begin{abstract}
Deep reinforcement learning (DRL) has demonstrated its potential in solving complex manufacturing decision-making problems, especially in a context where the system learns over time with actual operation in the absence of training data. One interesting and challenging application for such methods is the assembly sequence planning (ASP) problem. In this paper, we propose an approach to the implementation of DRL methods in ASP. The proposed approach introduces in the RL environment parametric actions to improve training time and sample efficiency and uses two different reward signals: (1) user\textsc{\char13}s preferences and (2) total assembly time duration. The user\textsc{\char13}s preferences signal addresses the difficulties and non-ergonomic properties of the assembly faced by the human and the total assembly time signal enforces the optimization of the assembly. Three of the most powerful deep RL methods were studied, Advantage Actor-Critic (A2C), Deep Q-Learning (DQN) and Rainbow, in two different scenarios: a stochastic and a deterministic one. Finally, the performance of the DRL algorithms was compared to tabular Q-Learning\textsc{\char13}s performance. After 10000 episodes the system achieved near optimal behaviour for the algorithms tabular Q-Learning, A2C and Rainbow. Though, for more complex scenarios the algorithm tabular Q-Learning is expected to underperform in comparison to the  other 2 algorithms. The results support the potential for the application of deep reinforcement learning in assembly sequence planning problems with human interaction.

\end{abstract}

keywords: Deep Reinforcement Learning, Manufacturing, Assembly Sequence Planning.

\section{Introduction}
\label{intro}
\begin{figure}[]
  \includegraphics[width=1\textwidth]{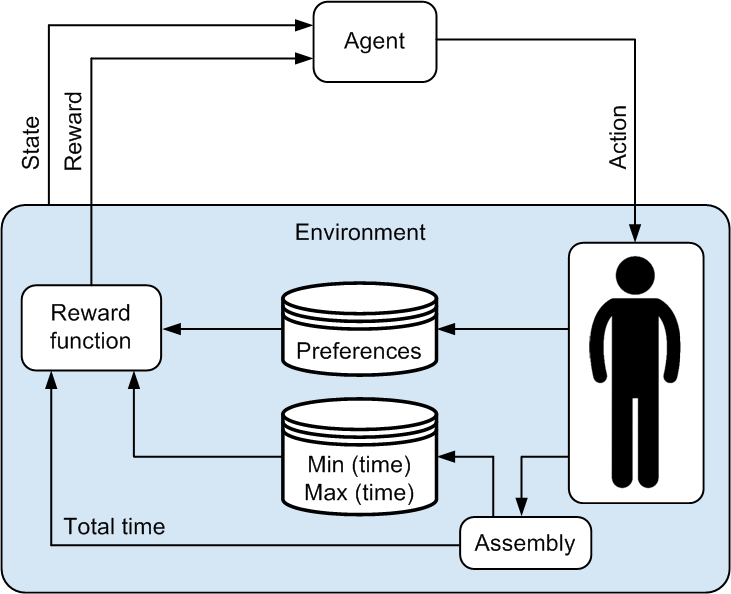}
\caption{Proposed method.}
\label{fig:1}       
\end{figure}

Reinforcement learning (RL) is a machine learning par\-a\-digm where an agent learns to solve a predefined task through continuous interactions with its environment. Multiple success stories have emerged from the implementation of reinforcement learning methods in game playing \cite{Ref1,Ref2,Ref3,Ref4} and robotic control tasks \cite{Ref5}. These success stories have encouraged the growth in the range of applications for reinforcement learning. With the advent of Industry 4.0 and its product customization ideology, there is a high demand for extremely efficient and optimized assembly processes, reason for the relatively high amount of machine learning studies in the optimization of production processes \cite{Ref6}. This raises the question on how viable and useful deep reinforcement learning methods would be on solving assembly sequence planning (ASP) optimization problems. This research intends to study the applicability of deep reinforcement learning (DRL) in ASP problems by introducing parametric actions in the RL environment, in order to improve training time and sample efficiency, and by using two different reward signals: (1) user\textsc{\char13}s preferences and (2) total assembly time duration, Fig.~\ref{fig:1}.

Multiple reinforcement learning algorithms were applied and compared to the ASP optimization problem of a toy airplane. Moreover, due to flexibility and complexity factors, multiple assembly processes are still performed by human operators. For this reason, the ergonomic properties and difficulty of the assembly sequences generated need to be taken into consideration since poorly designed assembly sequences may lead to a faster deteriorating performance of the human operator due to fatigue. Therefore, user\textsc{\char13}s preferences were introduced as an additional reward signal to the learning agent.
The demand for highly efficient assembly processes has led to an increased interest in Human-Robot Collaboration (HRC) due to the possibility of combining the robot\textsc{\char13}s repeatability, consistency, and efficiency with the human\textsc{\char13}s dexterity, flexibility, and adaptability. As a consequence, multiple studies have researched the usage of reinforcement learning in HRC contexts. Ghadirzadeh et al. devised a novel framework where the robot balances the benefits of taking a proactive action with the risk of taking an incorrect action, allowing for a more time-efficient collaboration between human and robot \cite{Ref7}. Handover has been addressed by recent studies as being one key skill of collaborative robots. Kshirsagar et al. applied the Guided Policy Search (GPS) method to a human-robot object handover task with moving targets, large variation in the target location and a variable end-effector mass \cite{Ref8}. This algorithm was shown to have a poor spatial generalizability over variations in the target position, which could be improved by the introduction of local controllers in regions with high errors. On the other hand, the learned policies were shown to generalize well over a decent range of end-effector masses. Reinforcement learning was also applied to a needle hand-off task in the context of suturing in Robot-Assisted Surgeries (RAS). The developed method was shown to be robust and consistent, allowing for a viable solution in the training of novice users of surgical robots \cite{Ref9}. Assembly processes are a common application of HRC due to the dexterity and adaptability required in many operations, which is unachievable by contemporary robotic systems. Oliff et al. developed a methodology capable of reducing idle time by improving the capability of the robotic operators to adapt to the human operator\textsc{\char13}s performance, without decreasing productivity levels \cite{Ref10}. In a similar setting, a method was developed to optimize the task sequence allocation in an assembly process by minimizing the total assembly time and difficulty \cite{Ref11}. The results achieved support the applicability of the devised method to the dynamic division of human-robot collaborative tasks. To optimize the completion time of a collaborative assembly, Yu et al. proposed the usage of a Markov game model, inspired on the chess game, as well as a novel algorithm named DQN-MARL. The proposed method and the Markov game model were proven to be effective in obtaining optimal task scheduling policies for complicated task structures \cite{Ref12}. An effective collaboration between a human and a robot is a challenging task due to the unpredictability of the human partner. As such, Buerkle et al. proposed an adaptive human sensor framework which receives objective, subjective and physiological metrics and was shown to be capable of predicting the perceived workloads on two different assembly settings \cite{Ref13}. This framework has also revealed a promising potential towards adaptive behaviour, which could lead to more effective HRC. Liu et. al. applied DQN to an environment containing two agents, one for the robot and the other one for the human, featuring tasks with dynamic and stochastic characteristics \cite{Ref14}. The results shown the proposed method is able to drive the robot agent to learn how to make decisions in a collaborative assembly task.
Reinforcement learning has also been successfully applied to robotic ASP problems. The usage of a cyber-physical assembly system for a fully-automated assembly sequence planning, which takes into account physical characteristics of the robotic arms, has been proven to be more efficient than the interactive methods most commonly applied in assembly sequence planning \cite{Ref15}. Watanabe and Inada developed a method to improve overall efficiency of the assembly process by generating the assembly sequence and assigning work to both robot hands of a dual-arm robot. It was also proposed a method to transfer past learning experience to new assemblies \cite{Ref16}. Mao et al. proposed a version of the DQN algorithm optimized by genetic algorithm (GA), GO-DQN, to be applied to virtual reality (VR) maintenance training, formalized as a disassembly sequence planning (DSP) problem \cite{Ref17}. This new method was shown to hold potential at on-site maintenance training. Additionally, job scheduling problems are ever more common in the manufacturing landscape due to the gradual replacement of traditional machines by robots. For this reason, Wang et al. proposed a multi-agent scheduling algorithm for job shop scheduling in resource preemption environments \cite{Ref18}. This multi-agent reinforcement learning algorithm is employed to learn both the decision-making policy of each agent and the cooperation between job agents. The proposed method was shown to outperform traditional rule-based methods and the distributed-agent reinforcement learning method. 
Another form of interaction between robot and human is the process of human guidance where the human provides guidance, i.e., some form of input, for the robotic system to learn. A wide range of frameworks for human guidance have been developed throughout the years and each framework has its own advantages \cite{Ref19}. Examples of such frameworks are the more standard Imitation Learning (IL), where the agent learns directly from human actions, Evaluative Feedback, where the human provides immediate feedback to the agent\textsc{\char13}s actions, Imitation from Observation, which is similar to the human demonstrated action, and Learning from Attention from Human, where the trainer provides an attention map to the agent. Zhan et al. proposed a method named GAN-assisted human preference-based reinforcement learning, which is based on preference-based reinforcement learning, an approach that aims to solve the impracticality of obtaining sufficient demonstrations by replacing human demonstrations by human queries. The new method extends on this approach through the usage of a generative adversarial network (GAN) to learn human preferences and then replaces the role of the human in assigning preferences \cite{Ref20}. The results have shown a reduction of approximately 99.8\% of human time while maintaining similar levels of performance. Kim et al. studied the applicability of error-related potentials, an event-related activity in the human electroencephalogram (EEG), as a guidance method. This approach was capable of correctly learning the mapping between gestures and actions in a gesture-based robot control setting.
The remainder of the article was divided in 5 chapters. Chapter~\ref{sec:2} addresses the proposed reinforcement learning framework. In chapter~\ref{sec:3} the deep reinforcement learning algorithms studied are described. Chapter~\ref{sec:4} introduces the case study and experimental setting. Chapter~\ref{sec:5} presents the results and chapter~\ref{sec:6} the conclusion and suggestions for future work.

\section{Background}
\label{sec:2}

Reinforcement learning problems are typically formalized as Markov Decision Processes (MDPs), which are defined by the tuple \(\left(S,A,P,r\right)\), where \textit{S} is the state space, \textit{A} is the action space, \(P:S\times A\times S\rightarrow \mathbb{R}\)  is the transition operator and \(r:S\times A\rightarrow \mathbb{R}\) is the reward function. The agent interacts with the environment and generates a trajectory sequence of states, actions, and rewards \(\tau:\left\{s_0,a_0,r_0,s_1,a_1,r_1,\ldots,s_T,a_T,r_T \right\}\), where \textit{T} is its length, by following a policy \(\pi:S\rightarrow \mathbb{R}\). The policy \textit{$\pi$} is a mapping from states to actions, i.e., an action is outputted for each state input. The discounted accumulated rewards \(R\left(\tau\right)\) are calculated through the following equation:

\begin{equation}\label{eq:1}
R\left(\tau\right)=\sum_{t=0}^{T}\gamma^{t}r_t\left(s_t,a_t\right),\; 0<\gamma<1
\end{equation}

The parameter \(\gamma \in \left[0,1\right]\) is a discount factor that determines the importance of future rewards in comparison with the immediate reward.
Reinforcement learning algorithms have the objective of maximizing the discounted accumulated reward and can be distinguished based on three main aspects.

\subsection{Policy-based vs value-based algorithms}
\label{sec:2_1}
In policy-based methods a policy \textit{$\pi$} that maps states to actions is learned directly through a process of gradient ascent. These methods are unbiased and have better convergence properties since they directly learn the policy. However, policy-based methods usually converge to local optimum, are sample-inefficient and have high variance. On the other hand, value-based methods learn value functions which are used to determine which actions to choose in each state. Value-based methods can be built to estimate value functions \(V^\pi \left(s_t\right)\), action-value functions \(Q^\pi \left(s_t,a_t\right)\), or advantage functions \(A^\pi \left(s_t,a_t\right)\), which are defined through the following equations, respectively, where \(\mathbb{E}\) is the expected value operator:

\begin{equation}\label{eq:2}
V^\pi\left(s_t\right)=\mathbb{E}_\pi\left[\sum_{k=t}^{T}\gamma^{k-t}\left(s_t,a_t\right)\mid s=s_t\right]
\end{equation}

\begin{equation}\label{eq:3}
Q^\pi\left(s_t,a_t\right)=\mathbb{E}_\pi\left[\sum_{k=t}^{T}\gamma^{k-t}\left(s_t,a_t\right)\mid s=s_t,a=a_t\right]
\end{equation}

\begin{equation}\label{eq:4}
A^\pi\left(s_t,a_t\right)=Q^\pi\left(s_t,a_t\right)-V^\pi\left(s_t\right)
\end{equation}

The value function corresponds to the expected accumulated reward starting from the state \(s_t\), the action-value function corresponds to the expected accumulated reward of taking the action \(a_t\) in the state \(s_t\), and the advantage function is the difference between the action-value function and the value-function, i.e., how does picking the action \(a_t\) compare to the average of all actions. Such algorithms can use samples from any behaviour policy, i.e., they are typically off-policy, which makes them more sample efficient than policy-based algorithms, which are typically on-policy algorithms. However, since value-based algorithms optimize for a surrogate Bellman error objective, they can be unstable, i.e., the convergence is not guaranteed.

\subsection{On-policy vs off-policy algorithms}
\label{sec:2_2}
On-policy algorithms use transition samples from the current policy to improve the policy while off-policy algorithms can reuse samples from any policy to make policy improvements. For this reason, off-policy algorithms are generally more sample efficient than on-policy algorithms.

\subsection{Model-free vs model-based algorithms}
\label{sec:2_3}
In model-based algorithms a predictive model of the environment is learned from the transition samples. This predictive model is then used to improve the policy. In the case of model-free algorithms the transition samples are used directly to learn a policy (policy-based algorithms) or to learn a function approximator (value-based algorithms).

\section{Methods and Methodologies}
\label{sec:3}

\subsection{Tabular Q-Learning}
\label{sec:3_1}

The tabular Q-Learning algorithm is based on the concept of learning a Q-table, which is a matrix that represents the Q-value for each state and action pair, i.e., a tabular representation of the state-action value function. The Q-table is updated after each step through the Bellman equation, where \(Q^{new} \left(s_t,a_t \right)\) is the updated Q-value for the state \(s_t\) and action \(a_t\), \(Q\left(s_t,a_t\right)\) is the previous Q-value, \textit{$\alpha$} is the learning rate, which controls how much the Q-value is updated and \(\max_a Q\left(s_{t+1},a\right)\) is an estimate of the maximum accumulated reward by taking the optimal action \textit{a} in the state \(s_{t+1}\):

\begin{equation}\label{eq:5}
\begin{split}
Q^{new}\left(s_t,a_t\right)\leftarrow{} & \left(1-\alpha\right)\times Q\left(s_t,a_t\right) \\
& + \alpha \left(r_t+\gamma \times \max_a Q\left(s_{t+1},a\right)\right)
\end{split}
\end{equation}

\subsection{A2C}
\label{sec:3_2}

The advantage actor-critic (A2C) algorithm has two distinct deep neural network (DNN) representations, the actor, and the critic \cite{Ref21}. An actor-critic algorithm is a hybrid algorithm since it incorporates both policy-based and value-based characteristics. The actor is grounded on policy-based leaning and has the purpose of deciding which actions to take and the critic is ground\-ed on value-based learning and has the purpose of evaluating the actions selected. A policy-based algorithm such as REINFORCE \cite{Ref22}, updates the parameters \textit{$\theta$} of the policy \(\pi\left(a\mid s\,;\,\theta\right)\) through the following gradient:

\begin{equation}\label{eq:6}
\nabla_\theta J\left(\theta\right)=\mathbb{E}_\tau\left[\sum_{t=0}^{T-1}\nabla_\theta \log\pi_\theta \left(a_t\mid s_t\right)R_t\right]
\end{equation}

The parameter \(\nabla_\theta\) is the gradient in respect to parameters \textit{$\theta$}, and \(J\left(\theta\right)\) is the expected sum of rewards under a trajectory distribution \(p_\theta\left(s_1,a_1,s_2,a_2,\ldots,s_T,a_T\right)=p_\theta \left(\tau\right)\). Updating the policy parameter through Monte Carlo updates, i.e. taking random samples, introduces high variance, leading to instability and slow convergence \cite{Ref22}. Therefore, to reduce the variance of the estimate, while keeping it unbiased, a learned function of the state, known as baseline, is subtracted from the cumulative reward \(R_t\). This baseline is represented by the critic network. The value function \(V^\pi \left(s_t\right)\), (\ref{eq:2}), is one commonly used baseline, and since \(R_t\) is an estimate of the state-action value function \(Q^\pi \left(a_t \mid s_t\right)\), by remembering (\ref{eq:4}), subtracting the value function from the state-action value function corresponds to the advantage function. The new gradient is displayed in the following equation:

\begin{equation}\label{eq:7}
\nabla_\theta J\left(\theta\right) = \mathbb{E}_\tau\left[\sum_{t=0}^{T-1}\nabla_\theta \log\pi_\theta\left(a_t\mid s_t\right)A_\varphi \left(s_t,a_t\right)\right]
\end{equation}

The baseline in an actor-critic algorithm is provided by the critic, with parameters \textit{$\varphi$}. The advantage estimate in the A2C algorithm is computed through the n-step advantage estimate:

\begin{equation}\label{eq:8}
A_\varphi\left(s_t,a_t\right)=\sum_{k=0}^{n-1}\gamma^k r_{t+k+1}+\gamma^n V_\varphi\left(s_{t+n+1}\right)-V_\varphi\left(s_t\right)
\end{equation}

Where \textit{r} is the reward, \textit{$\gamma$} the discount factor and \(V_\varphi\) is the value function estimated by the critic network. Finally, from (\ref{eq:7}) and (\ref{eq:8}) the actor objective function is obtained:

\begin{equation}\label{eq:9}
\begin{split}
\nabla_\theta J\left(\theta\right)= & \sum_t\nabla_\theta\log\pi_\theta\left(s_t,a_t\right) \\ & \times\Biggl[\sum_{k=0}^{n-1} \gamma^k r_{t+k+1} \\
& \quad +\gamma^n V_\varphi\left(s_{t+n+1}\right)-V_\varphi\left(s_t\right)\Biggr]
\end{split}
\end{equation}

On the other hand, the critic network is updated through the loss function displayed:

\begin{equation}\label{eq:10}
L\left(\varphi\right)=\sum_t\left(R_t-V_\varphi\left(s_t\right)\right)^2
\end{equation}

\subsection{DQN}
\label{sec:3_3}

The DQN algorithm is an off-policy value-based algorithm [1]. The optimal action-value function \(Q^* \left(s_t,a_t \right)\) is defined as the maximum expected accumulated reward achievable, starting from a state \(s_t\) and taking the action \(a_t\):

\begin{equation}\label{eq:11}
 Q^*\left(s_t,a_t\right)=\max_\pi\mathbb{E}\left[R_t\mid s=s_t,a=a_t,\pi\right]
\end{equation}

This optimal action-value function obeys the Bellman equation, which is based on the idea that, by knowing the optimal action-value function \(Q^* \left(s_t,a_t\right)\) for any state \(s_t\) and action \(a_t\), the optimal strategy would be to select the action \textit{a} that maximizes the expected value:

\begin{equation}\label{eq:12}
 Q^*\left(s_t,a_t\right)=\mathbb{E}_{s\textsc{\char13}\sim\mathcal{E}}\left[r_t+\gamma\max_a Q^*\left(s_{t+1},a\right)\mid s_t,a_t\right]
\end{equation}

This optimal action-value function is estimated by a neural network function approximator with weights \textit{$\theta$}, known as Q-network, \(Q\left(s_t,a_t;\theta\right)\approx Q^*\left(s_t,a_t\right)\). The Q-network is trained by minimizing a sequence of loss functions \(L_i \left(\theta_i\right)\) at each iteration \textit{i}:

\begin{equation}\label{eq:13}
L_i\left(\theta_i\right)=\mathbb{E}_{s_t,a_t\sim\rho\left(\cdot\right)}\left[\left(y_i-Q\left(s_T,a_t;\theta_i\right)\right)^2\right]
\end{equation}

Where \(\rho\left(s_t,a_t\right)\) is a probability distribution and \(y_i\) is the target at each iteration \textit{i}:

\begin{equation}\label{eq:14}
y_i=\mathbb{E}_{s_{t+1}\sim\mathcal{E}}\left[r_t+\gamma\max_a Q\left(s_{t+1},a;\theta_{i-1}\right)\mid s_t,a_t\right]
\end{equation}

During the loss function optimization \(L_i \left(\theta_i\right)\), the parameters from the previous iteration \(\theta_{i-1}\) are kept fixed. The gradient can then be obtained by differentiating the loss function in respect to the weights \(\theta_i\):

\begin{equation}\label{eq:15}
\begin{split}
\nabla_{\theta_i} L_i\left(\theta_i\right)= & \mathbb{E}_{s_t,a_t\sim\rho\left(\cdot\right);s_{t+1}\sim\mathcal{E}}\biggl[(r_t \\
 & \quad +\gamma\max_a Q\left(s_{t+1},a;\theta_{i-1}\right) \\
 & \quad -Q\left(s_t,a_t;\theta_i\right))\nabla_{\theta_i} Q\left(s_t,a_t;\theta_i\right)\biggr]
\end{split}
\end{equation}

This algorithm employs the experience replay technique, where transition samples at each time-step \textit{t}, \(e_t=\left(s_t,a_t,r_t,s_{t+1}\right)\), are stored in a dataset \(\mathcal{D}=e_1,\ldots,e_N\). Then, minibatch updates are applied to samples of experience, \(e\sim\mathcal{D}\), which are drawn at random from the pool of samples. 
This approach allows the reusability of samples, which leads to better sample-efficiencies. Also, by randomizing the samples, the correlations between consecutive samples are broken down and therefore the variance of the updates is reduced. Lastly, in an on-policy setting, the data samples are dependent on the current parameters, i.e., if a certain action maximizes the accumulated reward, the following training samples will correspond mostly to samples where this action was selected. If at a certain moment a different action is considered as best, the training distribution will also change. This behaviour can lead to either convergence to poor local optima or even divergence. Therefore, by the usage of the experience replay method the behaviour distribution is averaged over many previous states, which smooths out learning and avoids oscillations or divergences in the parameters.

\subsection{Rainbow}
\label{sec:3_4}

The Rainbow algorithm \cite{Ref23} combines multiple extensions proposed for the enhancement of DQN, namely Double DQN, prioritized experience replay, duelling networks, Distributional Q-Learning and Noisy DQN. Double DQN was proposed to address the overestimation bias of Q-Learning by decoupling selection and evaluation of the bootstrap action. Prioritized experience replay aims to improve sample efficiency by replaying transitions from which there is more to learn. Duelling network architecture was proposed to generalize across actions by representing state-values and action-values separately. Through the usage of multi-step bootstrap targets the bias-variance trade-off is shifted and newly observed rewards are propagated faster to earlier visited states. Distributional Q-Learning learns a categorical distribution of discounted rewards instead of estimating the mean. Noisy DQN uses stochastic network layers for exploration. The combination of these multiple improvements to DQN leads to a faster, more stable and more sample-efficient algorithm.

\section{Case Study}
\label{sec:4}

\begin{figure*}
  \includegraphics[width=1\textwidth]{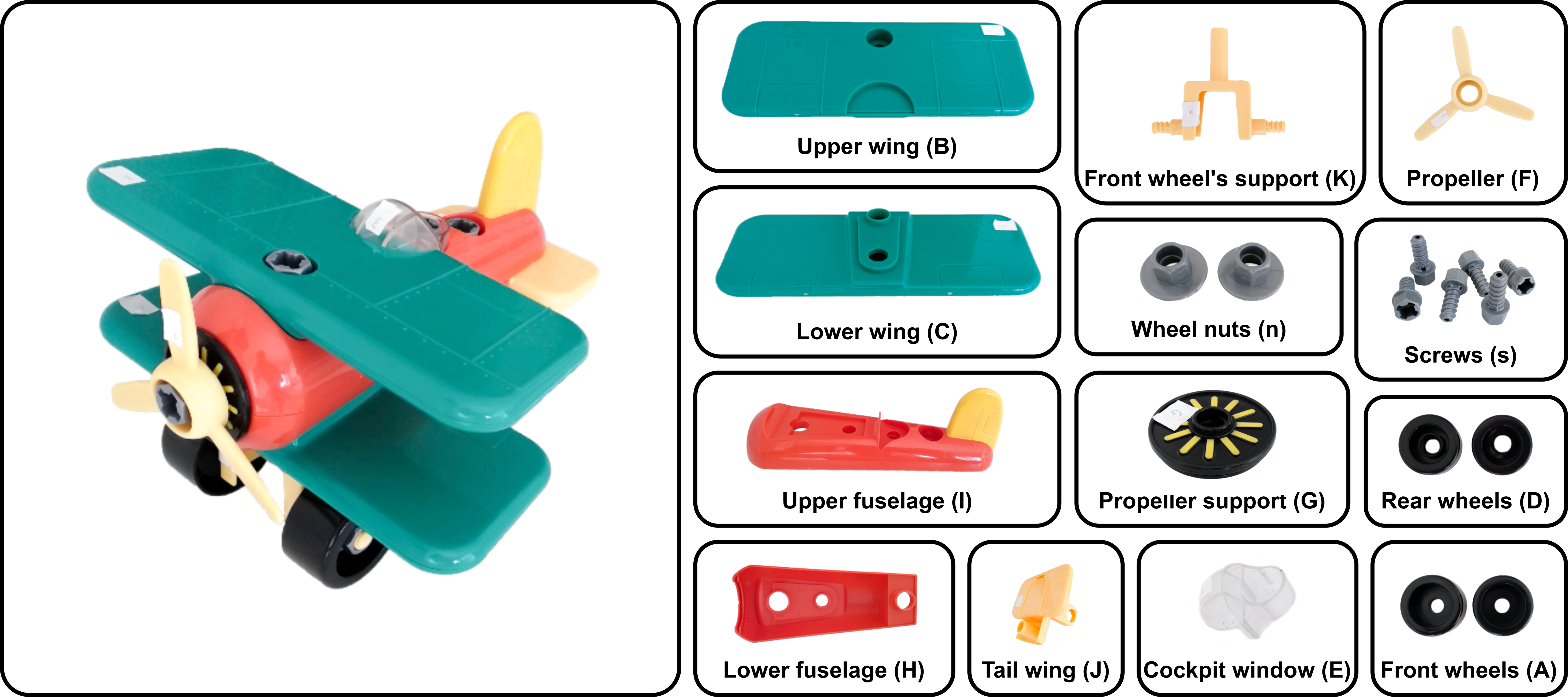}
\caption{Airplane toy from the Yale-CMU-Berkeley Object and Benchmark Dataset.}
\label{fig:2}       
\end{figure*}

The studied ASP problem was based on the assembly case study proposed by Neves et al. \cite{Ref24} of an airplane toy from the Yale-CMU-Berkeley Object and Benchmark Dataset \cite{Ref25,Ref26}, Fig~\ref{fig:2}. This assembly was optimized through the usage of the deep reinforcement learning algorithms A2C \cite{Ref21}, DQN \cite{Ref1}, and Rainbow \cite{Ref23}, provided in the RLlib python library \cite{Ref27}, and the tabular Q-Learning algorithm \cite{Ref28}.

\subsection{Assembly analysis}
\label{sec:4_1}

\begin{figure}[]
  \includegraphics[width=1\textwidth]{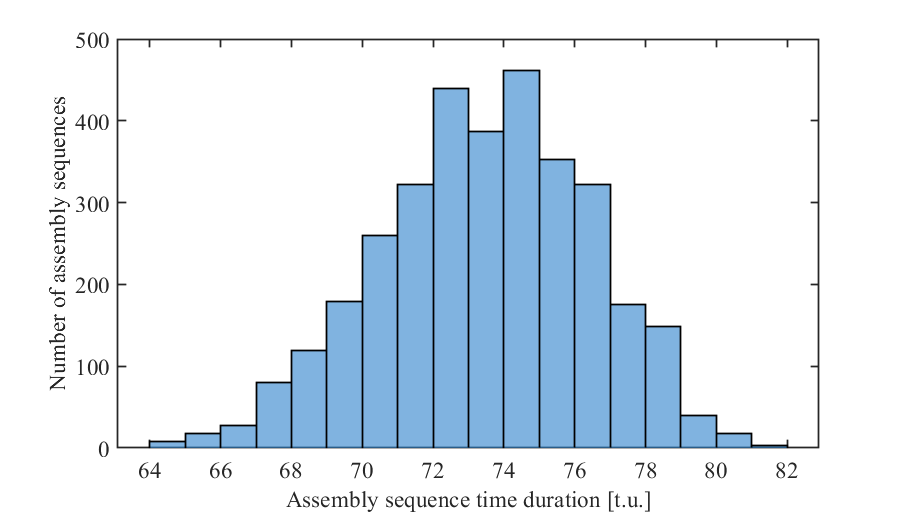}
\caption{Distribution of possible assembly sequences\textsc{\char13} time durations.}
\label{fig:3}       
\end{figure}

The airplane encompasses 11 distinct components, 2 types of fasteners (Table~\ref{tab:1}) and its assembly was subdivided into 8 elementary tasks. There are a total of 3360 possible assemblies, due to task dependencies (Table~\ref{tab:2}), with a mean assembly duration of 73.2 t.u. (time units), a maximum assembly duration of 82.0 t.u. and a minimum assembly duration of 64.0 t.u., following the assembly time duration distribution presented in Fig.~\ref{fig:3}.

\begin{table}[]
\caption{Airplane\textsc{\char13}s components and fasteners.}
\label{tab:1}       
\begin{tabular}{p{0.5cm}p{4.8cm}p{1.7cm}}
\hline\noalign{\smallskip}
Part & Aeroplane component\textsc{\char13}s description & Number of components  \\
\noalign{\smallskip}\hline\noalign{\smallskip}
A & Front wheels & 2 \\
B & Upper wing & 1 \\
C & Lower wing & 1 \\
D & Rear wheels & 2 \\
E & Cockpit window & 1 \\
F & Propeller & 1 \\
G & Propeller\textsc{\char13}s support & 1 \\
H & Lower body of the aeroplane & 1 \\
I & Upper body of the aeroplane & 1 \\
J & Rear body of the aeroplane & 1 \\
K & Front wheel\textsc{\char13}s support & 1 \\
\noalign{\smallskip}\hline\noalign{\smallskip}
n & Wheel nuts & 2 \\
s & Screws & 5 \\
\noalign{\smallskip}\hline
\end{tabular}
\end{table}

\begin{table}[]
\caption{Task precedence rules.}
\label{tab:2}       
\begin{tabular}{p{3.7cm}p{3.7cm}}
\hline\noalign{\smallskip}
Task & Precedence task  \\
\noalign{\smallskip}\hline\noalign{\smallskip}
1 & None \\
2 & 1 \\
3 & 1 \\
4 & 1 \\
5 & 1 and 4 \\
6 & 1 \\
7 & None \\
8 & None \\
\noalign{\smallskip}\hline
\end{tabular}
\end{table}

\subsection{MDP formulation}
\label{sec:4_2}

The MDP formulation of the airplane assembly sequence planning problem was based on the approach proposed by Neves et al. \cite{Ref24}.

\subsubsection{Observation}
\label{sec:4_2_1}

\begin{figure}[]
  \includegraphics[width=1\textwidth]{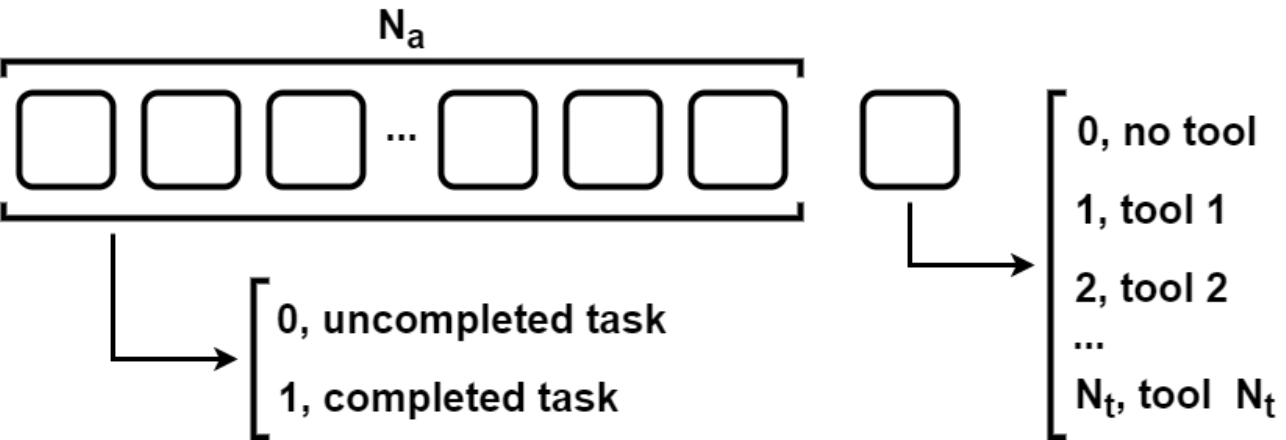}
\caption{State representation.}
\label{fig:4}       
\end{figure}

The observation is encoded by \(N_{a+1}\) digits, where \(N_a\) corresponds to the total number of tasks and the extra digit corresponds to the currently selected tool. The first \(N_a\) digits are binary so that if task \textit{k} was already executed the \(k^{th}\) digit has the value 1, otherwise it has the value 0. The digit related to the tool has values ranging from 0 (no tool selected) to the total number of unique tools \(N_t\). In this case study, there are a total of 2 different tools and the tool number 1 is required in all tasks except for the task 7, which requires the tool number 2. The proposed state representation is displayed in Fig.~\ref{fig:4}.

\subsubsection{Action}
\label{sec:4_2_2}

The actions selected by the agent are directly correlated to the assembly tasks, i.e., in this case study the agent has a total of 8 different available tasks to select from. Nevertheless, due to task dependencies, in every state some actions are restricted.

\subsubsection{Reward}
\label{sec:4_2_3}

The reward function considers two main criteria, the total assembly time minimization, and the user\textsc{\char13}s preferences. In a real-world setting the time duration of an assembly can only be known at the end of the episode, i.e., at the end of the assembly, and, as such, an episodic reward setting was considered. The user preference criterion allows the user to inform the agent of unwanted assembly sequences due to ergonomic or difficulty reasons. Regarding the total assembly time criterion, applying the time duration directly to the reward function may destabilize the performance of the DRL algorithms, and, as such, a normalization of the time duration was considered. The total assembly time is normalized between the minimum assembly time (\textit{minT}) and the maximum assembly time (\textit{maxT}) observed in previous episodes. To reduce the probability of outliers in the maximum assembly time value, i.e., extremely large assembly times due to some problem faced during the assembly, the values of assembly time are only considered to this value if it is within 2 standard deviations of the previous 100 episodes time durations. The resulting reward function is defined by:

\begin{equation}\label{eq:16}
r=\left\{\begin{array}{cl}
0 & \; \begin{array}{l}
if\; unfinished
\end{array}\\ 
\hphantom{} \\
0 & \; \begin{array}{l}
if\; first, \; not \; unwanted,\\ finished\; assembly
\end{array}\\ 
\hphantom{} \\
-1 & \; \begin{array}{l}
if\; finished\; and \; unwanted
\end{array}\\ 
\hphantom{} \\
\frac{maxT-T_a}{maxT-minT} & \; \begin{array}{l}
otherwise
\end{array}
\end{array}\right.
\end{equation}

The total assembly time duration \(T_a\) is obtained for 2 different settings through the following equation:

\begin{equation}\label{eq:17}
r=\left\{\begin{array}{cl}
\sum_{i=1}^{N_a}t_i & \; \begin{array}{l}
if\; deterministic
\end{array} \\
\sum_{i=1}^{N_a}\mathcal{N}\left(t_i,0.1 t_i\right) & \; \begin{array}{l}
if\; stochastic
\end{array}
\end{array}\right.
\end{equation}

The element \(t_i\) corresponds to the deterministic time duration of the task \textit{i}:

\begin{equation}\label{eq:18}
t_i=t_{b_i}+\sum_{k=1}^{N_a}y_k t_{c_{ki}}+x_i t_t
\end{equation}

Where \(t_{b_i}\) is the mean base time of the task \textit{i}, without any other previous tasks done, apart from the tasks required due to task dependencies (Table~\ref{tab:3}), \(t_{c_{ki}}\) are the time correction elements due to already done tasks (Table~\ref{tab:4}), where \(y_k\in\{0,1\}\) encodes whether the task \textit{k} has already been done, \(t_t\) is the tool change time, which has a value of 2 time units, and \(x_i\in\{0,1\}\) is a parameter that encodes the necessity of a tool change. The values for the task\textsc{\char13}s mean base time and time correction elements were obtained from the work done by Neves et al. \cite{Ref24}.

\begin{table*}[]
\caption{Task\textsc{\char13}s mean base time \(t_{b_i}\).}
\label{tab:3}       
\begin{tabular}{p{4.8cm}p{1.1cm}p{1.1cm}p{1.1cm}p{1.1cm}p{1.1cm}p{1.1cm}p{1.1cm}p{1.1cm}}
\hline\noalign{\smallskip}
Task & 1 & 2 & 3 & 4 & 5 & 6 & 7 & 8 \\
\noalign{\smallskip}\hline\noalign{\smallskip}
Average time [time units] & 10 & 7 & 8 & 6 & 12 & 8 & 11 & 9 \\
\noalign{\smallskip}\hline
\end{tabular}
\end{table*}

\begin{table}[]
\caption{Tasks\textsc{\char13} time correction elements \(t_{c_{ki}}\) [t.u.].}
\label{tab:4}       
\begin{tabular}{p{1.65cm}p{0.3cm}p{0.6cm}p{0.6cm}p{0.6cm}p{0.3cm}p{0.3cm}p{0.3cm}p{0.3cm}}
\hline\noalign{\smallskip}
Task & 1 & 2 & 3 & 4 & 5 & 6 & 7 & 8 \\
\noalign{\smallskip}\hline\noalign{\smallskip}
Task 1 done &  &  &  &  &  &  & 0 & 0 \\
Task 2 done &  &  & -1 & -1.5 & 0 & -1 & 0 & 1 \\
Task 3 done &  & 0 &  & 0 & 0 & 0 & 0 & 0 \\
Task 4 done &  & -0.5 & 0 &  &  & 0 & 0 & 0 \\
Task 5 done &  & -1 & -0.5 &  &  & -2 & 1 & 0 \\
Task 6 done &  & 0 & 0 & 0 & 0 &  & 0 & 0 \\
Task 7 done & 0 & 0 & 0 & 0 & 0 & 0 &  & 0 \\
Task 8 done & 0 & 0 & 0 & 0 & 0 & 0 & 0 &  \\
\noalign{\smallskip}\hline
\end{tabular}
\end{table}

In a deterministic setting the accumulated time is calculated by the sum of all tasks\textsc{\char13} deterministic time durations, which in turn is the sum of the task base time \(t_b\), the sum of all time correction elements due to already completed tasks \(\sum_{k=0}^ny_k t_{c_k}\), and the tool change time \(t_c\) if a tool change is required. In the setting where the accumulated assembly duration is stochastic, the duration of each task is obtained from a normal distribution with mean equal to the deterministic task assembly time and with standard deviation equal to 10\% of the deterministic task assembly time duration. The stochastic setting aims to, more closely, resemble a real-world problem.

\section{Results}
\label{sec:5}

\begin{figure}[]
  \includegraphics[width=1\textwidth]{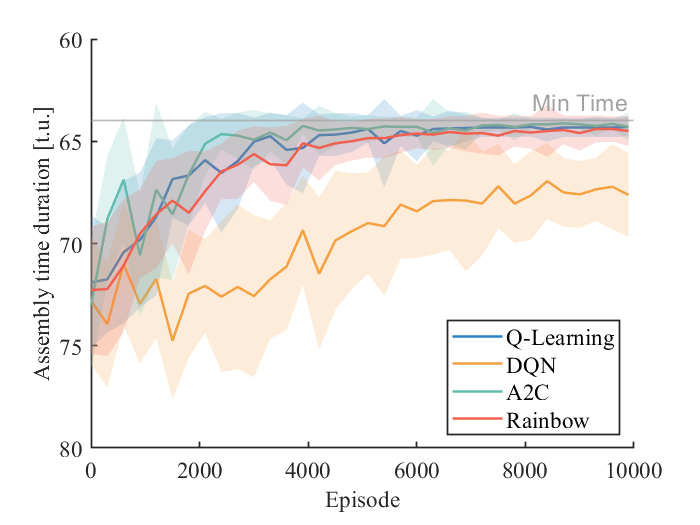}
\caption{Assembly time duration throughout the episodes in the deterministic setting.}
\label{fig:5}       
\end{figure}

\begin{figure}[]
  \includegraphics[width=1\textwidth]{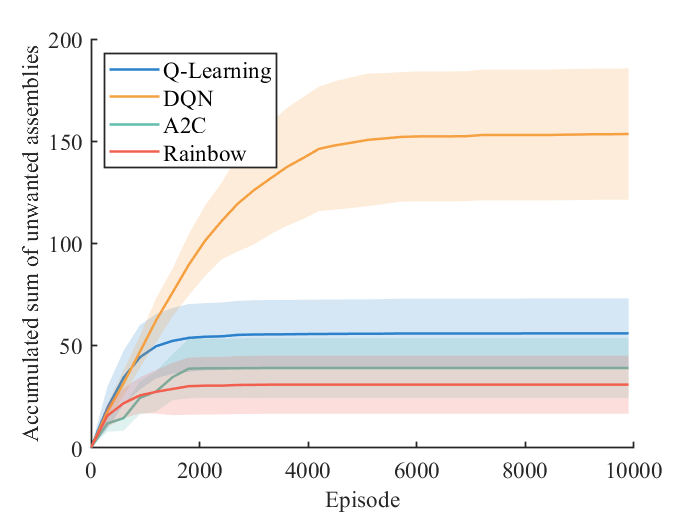}
\caption{Accumulated sum of unwanted assembly sequences in the deterministic setting.}
\label{fig:6}       
\end{figure}

As previously stated, the airplane assembly sequence planning problem was tested in two different settings, one with deterministic assembly time durations and one with stochastic assembly time durations. A total of 4 distinct algorithms were applied to the optimization problem (tabular Q-Learning, DQN, A2C and Rainbow). After an initial hyperparameter tuning the algorithm\textsc{\char13}s performance was analysed over 20 different trials with different seeds for both settings \cite{Ref29}. All algorithms, apart from Rainbow, were used with parametric actions, i.e., the agent was only allowed to select possible actions, based on task dependencies. This option was not applied in the Rainbow algorithm as it was not available for this algorithm. In Fig.~\ref{fig:5} the achieved assembly time durations throughout the learning episodes is displayed for the deterministic setting. As it can be observed the DQN algorithm had the worst performance. Effectively, this algorithm was the slowest to learn, reached a suboptimal mean assembly time of 67.6 ± 2.0 time units and had a large variability in results. The algorithms tabular Q-Learning, A2C and Rainbow shared a similar learning time. The tabular Q-Learning, A2C and Rainbow algorithms achieved a mean assembly time of 64.3 ± 0.5 t.u., 64.3 ± 0.7 t.u. and 64.5 ± 0.7 t.u., respectively. One possible reason for the higher variability in results and slower stabilization in the Rainbow algorithm is the lack of action restrictions, which requires the algorithm to search over a larger action space than other algorithms. Fig.~\ref{fig:6} displays the accumulated sum of unwanted trajectories throughout the learning phase. As in the assembly time performance, DQN presents the worst performance in respect to the occurrence of unwanted assembly sequences, reaching a mean of approximately 154 ± 32 occurrences. The algorithms tabular Q-Learning, A2C and Rainbow had a similar performance, reaching a mean occurrence of unwanted assemblies of 56 ± 17, 39 ± 15 and 31 ± 14 respectively. Nevertheless, the tabular Q-Learning algorithm had a slightly higher mean occurrence of unwanted assemblies.

\begin{figure}[]
  \includegraphics[width=1\textwidth]{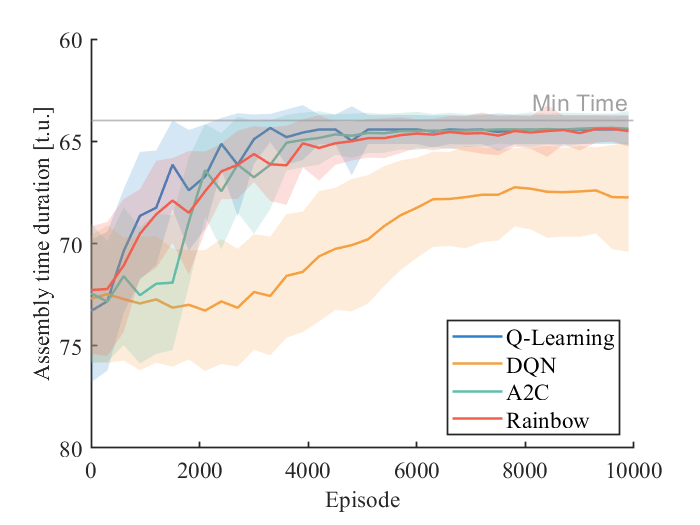}
\caption{Assembly time duration throughout the episodes in the stochastic setting.}
\label{fig:7}       
\end{figure}

\begin{figure}[]
  \includegraphics[width=1\textwidth]{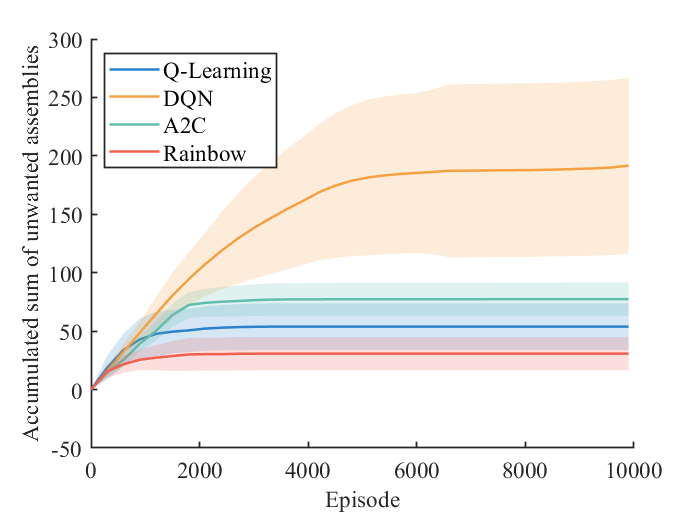}
\caption{Accumulated sum of unwanted assembly sequences in the stochastic setting.}
\label{fig:8}       
\end{figure}

Similarly, for the stochastic setting, Fig.~\ref{fig:7} and Fig.~\ref{fig:8} present the assembly time duration and the cumulative sum of unwanted assemblies throughout the learning phase for all 4 algorithms, respectively. For a more straightforward comparison, the values presented in Fig.~\ref{fig:8} correspond to the assembly time durations that would have been obtained in the deterministic setting instead of the assembly time durations experienced. Therefore, the variabilities observed only depend on the variability of the algorithms\textsc{\char13} performance in the stochastic scenario. As in the deterministic setting, DQN is the poorest performing algorithm for both criteria. It has the slowest learning, the worst achieved result for the mean assembly time duration, 67.7 ± 2.7 t.u., and the highest cumulative sum of unwanted assemblies, with a mean of 192 ± 75. As in the deterministic setting the algorithms tabular Q-Learning, A2C and Rainbow had a similar performance in the stochastic setting, with a mean achieved assembly time duration of 64.4 ± 0.7 t.u., 64.4 ± 0.9 t.u. and 64.5 ± 0.7 t.u. respectively. The accumulated sum of unwanted assemblies was 54 ± 20, 77 ± 14 and 31 ± 14 for the algorithms tabular Q-Learning, A2C and Rainbow respectively. Once again, the tabular Q-Learning algorithm had a slightly higher mean occurrence of unwanted assemblies.

Even though the tabular Q-Learning achieved a similar performance to the A2C and Rainbow algorithms, it is known to suffer from the “curse of dimensionality”, which refers to the exponential increase in the data and computation required to cover the entire state-action space in high-dimensional spaces, when the number of dimensions grows. As a consequence, by increasing the assembly complexity, i.e., more tasks, components, and tools, it should be expected a worse performance of this algorithm in comparison to the A2C and Rainbow algorithms.

\section{Conclusion and future work}
\label{sec:6}

A total of 4 reinforcement learning algorithms, tabular Q-Learning, DQN, A2C and Rainbow, were applied to the assembly sequence planning optimization problem of a toy airplane. Two scenarios were studied, one with deterministic assembly time durations and one with stochastic time durations. In both scenarios the DQN algorithm presented the worst performance, reaching suboptimal assembly sequence time durations and experiencing a 4 to 5 times higher number of unwanted assembly sequences when compared to the other 3 algorithms. The algorithms tabular Q-Learning, A2C and Rainbow had similar performances and were able to achieve near optimal assembly times in both deterministic and stochastic scenarios. One important factor to consider is the larger action space in the Rainbow algorithm, since one of the limitations of this research was not having the option of parametric actions for this algorithm. This factor predictably would lead to a slower learning and a higher variability. As such, it would be interesting to consider, in future work, the study of a similar scenario with a Rainbow algorithm with parametric actions. The algorithm Q-Learning is known to suffer from the “curse of dimensionality”. Therefore, by increasing the assembly complexity, it should be expected a worse performance of this algorithm in comparison to the A2C and Rainbow algorithms. As a result, in future work, more complex assembly processes should be considered and studied.
It can also be concluded that assembly sequence planning problems with human interaction are a possible application for deep reinforcement learning since, after approximately 10000 episodes the system was capable of learning near optimal behaviour while avoiding unwanted assemblies. Though, the sample efficiency should be improved to achieve a more effective and efficient approach.

\section{Declarations}

\subsection{Ethical Approval}
Not applicable

\subsection{Consent to Participate}
Not applicable

\subsection{Consent to Publish}
Not applicable

\subsection{Authors Contributions}
Miguel Neves implemented the methods and conducted the testing. Pedro Neto defined the initial approach and managed the experimental tests.

\subsection{Funding}
This research was partially supported by project PRODUTECH4S\&C (46102) by UE/FEDER through the program COMPETE 2020 and the Portuguese Foundation for Science and Technology (FCT): COBOTIS (PTDC/EME-EME/32595/2017) and UIDB/00285/2020.

\subsection{Competing Interests}
There are no competing interests, whether financial or non-financial.

\subsection{Availability of data and materials}
There are no data available, other than those reported in the document.

\subsection{Code availability}
There is no code available.




\end{document}